**Large Language Models and the Rationalist-Empiricist Debate.**

**By**

Dr David King,

Research Associate.

University of Glasgow.

School of Humanities.

David.A.King@glasgow.ac.uk

**Abstract.**

To many Chomsky's debates with Quine and Skinner are an updated version of the Rationalist-Empiricist debates in the 17$^{th}$ century. The consensus being that Chomsky's rationalism was victorious. This dispute has reemerged with the advent of Large Language Models. With some arguing that LLMs vindicate rationalism because of the necessity of building in innate biases to make them work. The necessity of building in innate biases is taken to prove that empiricism hasn't got the conceptual resources to explain linguistic competence. Such claims depend on the nature of the empiricism one is endorsing. Externalized empiricism has no difficulties with innate apparatus once they are determined empirically (Quine 1969). Thus, externalized empiricism isn't refuted because of the need to build in innate biases in LLMs. Furthermore, the relevance of LLMs for the rationalist-empiricist debate in relation to humans is dubious. For any claim about whether LLMs learn in an empiricist manner to be relevant to humans it needs to be shown that LLMs and humans learn in the same way. Two key features distinguish humans and LLMs. Humans learn despite a poverty of stimulus and LLMs learn because of an incredibly rich stimulus. Human's linguistic outputs are grounded in sensory experience and LLMs are not. These differences in how the two learn indicates that they both use different underlying competencies to produce their output. Therefore, any claims about whether LLMs learn in an empiricist manner are not relevant to whether humans learn in an empiricist manner.

**Introduction.**

This paper will evaluate the rationalist-empiricist debate in relation to Large Language Models[1]. Theorists have argued that because LLMs such as ChatGPT demonstrate surprisingly effective linguistic output, they provide evidence about human linguistic capacities (Piantadosi, 2023). It has also been suggested that LLMs are data which can be used to adjudicate between rationalist-empiricist debate (Buckner, C. 2018, Childers et al 2023, Long 2024). The paper will first consider the recent historical debate between rationalists and empiricists centring on the dispute between Chomsky, Skinner and Quine. This historical debate will illustrate what was traditionally at stake in the dispute. Once this is done the paper will evaluate whether the analogy between LLMs and human linguistic output is a good one; demonstrating that it is not an accurate analogy. And hence LLMs have little to tell us about the rationalist/empiricist debate pertaining to human cognitive capacities.

From about 1950 there was a resurgence of interest in the Rationalist/Empiricist debate. With people viewing Chomsky as an updated rationalist carrying on the traditions of Leibniz and Descartes. While Quine and Skinner were viewed as modern day empiricists carrying on the traditions of Locke and Hume. And a consensus emerged that Chomsky's rationalism won out over the empiricism of Quine and Skinner (Lyons1970, Hornstein 2005, Boeckx 2006, Laka 2009). In recent years with the rise of LLMs some have argued that their architecture is data driven and hence they are an existence proof that empiricist learning is a viable way of modelling the mind (Milliere & Buckner. 2024a). Prompting debate where others argue that these models don't vindicate any kind of empiricism as they rely on innate architecture (Childers et al 2023).

It will be demonstrated that LLMs are too dissimilar to human linguistic capacities to be used as a model for them. LLMs acquire their linguistic ability through being trained on linguistic data orders of magnitude higher than what humans require to acquire their native language (Milliere, R & Buckner C. 2024a). Furthermore, while human language is sensorily grounded, LLMs are not (Harnard 2024). These key differences demonstrate that LLMS and humans learn through different underlying competencies therefore any claims as to whether LLMs learn in an empiricist manner has no significance for how humans learn.

In the final section it will be argued that given that human cognition is currently the best example on the planet of general intelligence, if we want to build AGI, then it would be appropriate to try to model them on human cognitive systems. But even here given that externalized empiricists have no difficulty with using innate architecture if it can be determined experimentally (Quine 1968). There seems to be less and less use for the distinction between architecture proposed by rationalists and empiricists. Any attempt to model an AGI system on human cognitive capacities should stick closely to the capacities postulated by our best science and avoid getting bogged down in arcane philosophical debates on empiricism vs rationalism.

**Chomsky and the Rationalist-Empiricist Debate.**

---

[1] Henceforth Large Language Models will be referred to as LLMs.

## Chomsky & Skinner.

Famously the rationalist-empiricist debate of the seventeenth century was revived in the 1960s. Noam Chomsky's 1959 review of Skinner's book Verbal Behaviour was the beginning of this revival. His review was critical of Skinner's empiricist model of Verbal Behaviour. Chomsky entitled his 1966 book 'Cartesian Linguistics: A Chapter in the History of Rationalist Thought', thus setting himself up as the heir apparent to Descartes rationalism. While Skinner was viewed by many as the heir to the mantle of the empiricist kingdom[2].

In Skinner's Verbal Behaviour he divided language up into seven Verbal Operants which he argued are controlled using his three-term contingency of antecedent behaviour and consequence. Chomsky criticized Skinner for taking concepts from the laboratory based on experiments with rats and pigeons and extending them into areas where there was no such experimental evidence[3]. He charged Skinner of either using the concepts literally in which case his views were false, or metaphorically in which case the concepts offered no more insight than our own vague concepts derived from folk-psychological intentional discourse (Skinner 1959).

In his 1965 book 'Aspects of a Theory of Syntax' Chomsky made his famous distinction between competence and performance. Chomsky argued that the only substantive theory of performance which would be possible would come through an understanding of underlying competence (Chomsky, 1965). And he illustrated this point through showing how elements from our underlying grammatical competence could predict and explain aspects of our linguistic performance. Since behaviourism was obviously primarily concerned with behaviour many viewed Chomsky's competence performance distinction as an attack on behaviourism. Many scientists influenced by Chomsky argued that behaviourism not having a competence-performance distinction meant that it couldn't be taken seriously as a science (Jackendoff 2002, Collins 2007).

Chomsky (1972, 1986) poverty of stimulus argument was deemed a further dent in the behaviourist project (Hornstein 2005, Lyons 1970). Chomsky used the structure dependence of language, exemplified through auxiliary inversion, as an example of syntactic knowledge which a person acquired despite a poverty of stimulus. He argued that a person could go through much or all their life without ever encountering evidence for the construction. The reasoning being that if the child learned the construction despite never encountering evidence for its structure. And the child didn't engage in trial-and-error learning where he tried out incorrect constructions which were systematically corrected by his peers until he arrived at the correct one (Crain and Nakayama 1987,

---

[2] While Skinner's project involved an empiricist epistemology. He wasn't heavily influenced by the ideas of the empiricist philosophers. Earlier Behaviourists such as Hull explicitly tried to model Hume's associationism on Pavlov's classical conditioning. See Smith 1986 pp. 231-232. For more recent attempts to update Hume using modern behavioural science see Wilson et al 2001.

[3] In the 70 years since Skinner wrote Verbal Behaviour a lot of experimental work has been done with human subjects on his verbal operants. Petursdottir & Baily 2017 p 223 noted that between 2004-2016 there were 369 empirical studies done on Skinner's Verbal Operants. While Jennings et al (2021) p. 97 noted that there were 72 experiments done on Skinner's operant the Intraverbal between 2015 and 2020. Indicating that in the years since Chomsky wrote his review it is no longer true that we only have evidence from studies on non-human animals.

Brown and Hanlon 1970). Then knowledge of the construction must be built into the child innately.

Theorists viewed poverty of stimulus arguments as further evidence that the behaviourist project was doomed to failure (Hornstein 2005). With many contrasting Chomsky's emphasis on innate knowledge with Skinner's supposed blank-slate philosophy (Pinker 2002). Thereby situating Skinner as a modern-day Locke battling with a modern-day Descartes (Chomsky), and the consensus was that modern science had shown that the rationalist position was the correct one (Boeckx 2006, Laka 2009).

### Quine and Chomsky.

The debate between Chomsky and Skinner was primarily focused on issues in linguistics, and psychology. In philosophy the rationalist-empiricist debate played out in a debate between Quine and Chomsky. Quine billed himself as an externalized empiricist whose primary aim was to explain how humans go from stimulus to science in a naturalistic manner. His entire project centred on naturalising both epistemology and metaphysics. On the epistemological side of things his need to explain how we go from stimulus to science would involve psychological speculations on how we acquire our language, and how we develop the ability to refer to objects. Quine was explicit in these speculations that any linguistic theory was bound to be behaviourist in tone since we acquire our language through intersubjective mouthing of words in public settings (Quine 1960). This commitment to behaviourism set Quine at odds with Chomsky.

In 1969 Chomsky's wrote a criticism of Quine called 'Quine's Empirical Assumptions'. This criticism noted that Quine's notion of a pre-linguistic quality space wasn't sufficient to account for language acquisition. That Quine's Indeterminacy of Translation Argument was trivial and amounted to nothing more than ordinary Underdetermination. And that Quine's invocation of the notion of the probability of a sentence being spoken was meaningless.

Skinner never replied to Chomsky[4] arguing that Chomsky so badly misunderstood his position that further dialogue was pointless. But Quine (1968) did reply; in his reply he charged Chomsky with misunderstanding his position and of attacking a strawman. On the issue of a prelinguistic quality space he argued that it was postulated as a *necessary* condition of acquiring the ability to learn from induction or reinforcement; he never thought it was a *sufficient* condition of our acquiring language (ibid). Quine argued that "the behaviourist was knowingly and cheerfully up to his neck in innate apparatus" (Quine 1976 p. 57). He further argued that indeterminacy of translation was additional to underdetermination and revealed difficulties with linguists and philosophers' uncritical usage of meanings, ideas and propositions. And finally, he noted that Chomsky misunderstood Quine's discussion of the probability of a sentence being spoken. Quine wasn't speaking about the absolute probability of a sentence being spoken, rather he was concerned with the probability of a sentence being spoken in response to queries in an experimental setting (Ibid).

This led to a series of back-and-forth exchanges between Chomsky and Quine. In his (1970) 'Methodological Reflections on Current Linguistic Theory', Quine criticized

---

[4] For an interesting reply on Skinner's behalf see Kenneth McCorquodale (1970).

Chomsky's notion of implicit rule following. Quine noted that there are two senses of rule-following he could make sense of (1) Being guided by a rule: A person following a rule they can explicitly state, (2) Fitting a Rule: A person's behaviour can conform to any of an infinite number of extensionally equivalent rules. But Quine charged Chomsky of appealing to a third type of rule (3) A rule that the person cannot state, but is nonetheless implicitly following, and this rule is a particular rule distinct from all the other extensionally equivalent rules that the persons behaviour conforms to. Chomsky (1975) correctly responded that Quine was again arbitrarily assuming that underdetermination was somehow terminal in linguistics but harmless in physics. As Chomsky approach became less about rules and more to do with parameters switching Quine's rule-following critique had less and less traction.

Chomsky's critique of Skinner achieved the status of almost a creation myth in cognitive science. With most introductory texts in psychology or cognitive science attributing Chomsky's review of Verbal Behaviour as being the death-knell of behaviourism and the birth of cognitive science. Whereas Chomsky's criticism of Quine wasn't as well known, and it had a more nuanced reading. While a lot of people came to the view that Chomsky won the debate; it didn't attain the creation myth status that the review had. A strongly Quinean influence still pervades the philosophy of language[5]. But in scientific circles neither Skinner nor Quines's views are considered scientifically respectable alternatives to Chomsky's linguistics (Hornstein 2005).

Outside of the realm of academic debates, in the popular press when Skinner is spoken about, he is referred to a blank slate theorist (Pinker 2002). Skinner, and Quine, are considered exemplars of the empiricist tradition and modern-day inheritors of John Locke's mantal. Chomsky, a self-described exemplar of the Cartesian Tradition, is viewed as a modern-day Descartes. The scientific consensus is that Chomsky's rationalism has won out over Quine and Skinner's empiricism (Laka 2009, Hornstein 2005, Boeckx 2006).

**Artificial Intelligence and Empiricism.**

In recent years with AI getting more and more sophisticated; philosophers, psychologists, and linguists have begun to explore what these AI systems tell us about the rationalist-empiricist debate. With some theorists arguing that empiricist architecture is responsible for the success of recent AI systems (Buckner 2018, Long, 2024). While others have argued that because the AI's architecture require substantial in-built biases, they in fact support rationalism (Childers et al 2023).

Buckner (2018) argued that Deep Convolutional Neural Networks[6] are useful models of mammalian cognition. And he further argued that these DCNNs use of "transformational abstraction", vindicated Hume's empiricist conception of how humans acquire abstract ideas. Childers et al (2023) have hit back at this view and have argued both that LLMs and DCNNs require built in biases for them to be successful. And they further argue that the need for built in biases in AI is analogous to the way Quine needed to posit innate apparatus to explain language acquisition. The need for Quine and AI systems to use innate apparatus, according to Childers et al, undermines any proposed empiricist model (Childers et al 2023 p. 72).

---

[5] In his "Quine and Chomsky and the Ins and outs of language" Barry Smith traces Quines influence on the works of McDowell, Davidson, Dummett, and Lewis. Arguing that Quine's influence in this respect is unfortunate because it obscures the contemporary science of linguistics derived from Chomsky's foundational work.

[6] Henceforth Deep Convolutional Neural Networks are referred to as DCNNs.

Childers et all's reading of the rationalist empiricist debate is extremely idiosyncratic. Their assertion that the postulation of any innate dispositions is an immediate weakening of empiricism is bizarre (Ibid p. 84). This reading of the rationalist-empiricist dispute doesn't stand up to scrutiny. Hume, and early arch-empiricist needed innate formation principles in the human mind to account for how we combine the ideas we receive from impressions into complex thoughts (Fodor 2003). And even Chomsky who is viewed as a paradigm exemplar of the rationalist tradition argued that innateness wasn't the issue when it came to the rationalist-empiricist debate:

> *"…Each postulates innate dispositions, inclinations, and natural potentialities. The two approaches differ in what they take them to be…The crucial question is not whether there are innate potentialities or innate structure. No rational person denies this, nor has the question been at issue. The crucial question is whether this structure is of the character of E or R; whether it is of the character of "powers or "dispositions"; whether it is a passive system of incremental data processing, habit formation, and induction, or an "active" system which is the source of "linguistic competence" as well as other systems of knowledge and belief" (Chomsky 1975 pp. 215-216)*

And Watson, Quine and Skinner were consistent about this point throughout their careers: Wason 1924 p. 135, Skinner 1953 p. 90, Skinner 1966 p. 1205, Quine 1969 p. 57, Quine 1973 p. 13, Skinner 1974, p.43.

The point of Childers et all's criticism was that Hume's empiricism with its appeal to a few laws of association., needed to be supplanted by Kant's system which postulated many more innate priors (Childers et al 2023 p. 87). This may have been a problem of Hume, but it is no difficulty for the likes of Quine who was an externalized empiricist who had no issues whatsoever with innate priors once they could be determined experimentally (Quine 1969 p. 57). When it comes to Artificial Intelligence there is a legitimate debate on whether it is pragmatic to build the systems on rationalist or empiricists principles. But this only has relevance to the rationalist empiricist debate if it can be demonstrated that artificial intelligence systems learn in the same way as humans do. In the next section I will evaluate how closely AI systems model human cognition. To do this I will focus LLMs and the degree to which they accurately model human linguistic cognition.

**Large Language Models and Human Linguistic Competence.**

Theorists have argued that the similarities between LLMs output and human linguistic output make LLMs and the way they learn directly relevant to theoretical linguistics. Thus, Piantadosi (2023), has argued that LLMs refute central claims made by Chomsky et al in the generative grammar tradition about language acquisition. This comparisons of LLMs to actual human cognition has been challenged in the literature (Chomsky et al 2023, Kodner et al 2024, Katzir, R 2023). In this section I will consider various disanalogies between LLMs and Human linguistic cognition which makes any comparison between them problematic. And in the final section I will consider the relevance of these disanalogies towards considering work in AI as being pertinent to debates about Rationalism versus Empiricism.

**Poverty of Stimulus Arguments, Artificial Intelligence, & Human Linguistic Capacities.**

A clear disanalogy to human linguistic abilities and LLMs is that humans acquire their language *despite* a poverty of stimulus, while LLMs learn *because* of a richness of stimulus (Kodner et al 2023, Chomsky et al 2023, Long, R 2024, Marcus, G. 2020). To see the importance of this distinction a brief discussion of the role that Poverty of Stimulus Arguments have played in linguistics is necessary, and with this in place we can return to the stimulus which LLMs are trained on.

Chomsky 1965 noted that people acquire syntactic knowledge despite a poverty of stimulus. Humans are exposed to limited fragmentary data and still manage to arrive at a steady state of linguistic competence, including knowledge of syntactic rules which they may not have ever encountered in their primary linguistic data. Chomsky used auxiliary inversion as his paradigm example of a poverty of stimulus (Chomsky 1965, 1968, 1971, 1972, 1975, 1986, 1988[7]). Pullum and Scholz (2002) reconstructed Chomsky's Poverty of Stimulus Argument as follows:

1. Humans learn their language either through data driven learning or innately primed learning.
2. If humans acquire their first language through data driven learning, then they can never acquire anything for which they lack crucial evidence.
3. But Infants do indeed learn things for which they lack crucial knowledge.
4. Thus, humans do not learn their first language by means of data-driven learning.
5. Conclusion: humans learn their first language by means of innately primed learning (Pullum and Scholz 2002).

Pullum & Scholz (2002) Isolated premise three as the key premise in the argument. And they sought empirical evidence to discover the amount of times constructions with evidence for auxiliary occur in a sample of written material. The material they choose to examine was Wall Street Journal back issues. They also estimated the amount of linguistic data a person is on average exposed to do. To do this they relied on Hart and Risely (1997) 'Meaningful Differences in the Everyday Experiences of Young Children'. They estimated that your average child from a middle-class background will have been exposed to about 30million word tokens by the age of three. Pullum and Scholz argue that the child will have been exposed to about 7500 relevant examples in three years. Which amounts to about 7 relevant questions per day. But a primary criticism of their work was that the Wall Street Journal wasn't representative of the type of data that a child would be exposed to. Sampson 2002 searched the British National Corpus which included child parent interactions and argued that the child would be exposed to about 1 relevant example every 10 days.

But the next question was whether a child would be able to learn the relevant construction from 1 example every 10 days (Lappin & Clark 2011). Reali and Christensen (2005) Perfors, Tenebaum & Reiger (2006) have all constructed mathematical models demonstrating that children are capable of learning from the above amounts of data. However, Berwick & Chomsky et al. (2011) in their 'Poverty of Stimulus Revisited' have hit back arguing that Auxiliary Inversion is meant as an expository example to illustrate the APS to the general public. And that there are much deeper syntactic properties which children could not learn from the PLD. The debate still rages on, but it is still a consensus in generative grammar that the Poverty of Stimulus is a real phenomenon which humans need domain specific innate knowledge to overcome.

---

[7] In the following I am Pullum and Scholz (2002) 'Empirical Assessment of Stimulus Poverty Arguments.

As our discussion above indicates that there has been some push back against Poverty of Stimulus Arguments however it is still the default position in linguistics. Furthermore, even those who push back against the APS would gleefully admit that the linguistic data a LLM is trained on is not remotely analogous to the Primary Linguistic Data of your average child. Children are exposed to 10million tokens a year, LLMs are exposed to around 300 billion tokens and this number is increasing exponentially (Kodner et al 2024). So, while the output of a LLM and a human may be roughly analogous the linguistic input they receive is in no way analogous.

### Competence & Performance in LLMs & Humans.

The divergence on linguistic data which LLMs and humans are trained on is a clear indicator that they work of different underlying competencies. Other differences emerge in terms of the materials they use. In terms of computation considerations chips are faster than neurons (Long 2024). To the degree that outputs are similar that doesn't demonstrate that they are implementing the same underlying competencies (Firestone 2020, Kodner et al 2024, Milliere & Buckner 2024b). Kodner et al give the example of two watches both of which keep time accurately but one of which is digital, and the other is mechanical. Despite similar performances they achieve it through different underlying competencies (Kodner et al 2024).

### But Are Human and LLM Outputs Analogous?

The question of whether Humans and LLM's output are analogous is obviously vital if we want to understand whether they operate using the same underlying competencies. We have already seen that the two systems seem to learn differently one *despite* the poverty of stimulus and one *because* of the richness of stimulus. This points towards different underlying innate competencies. Different underlying competencies aside the next section will demonstrate that the performances of each system are very different.

At a superficial level it LLMs and Human outputs appear very similar. Chat GPT can to some degree fool a competent reader into thinking that a human produced the outputted sentences. Clark et al (2021) studied human created stories, news articles, and texts and got LLMs (GPT 2 and GPT 3) to create similarly sized stories. 130 participants were tested, and 50% of the participants couldn't tell the human written text apart from the LLM models outputs. (Scwitzgebel et al 2023). Scwitzgebel, et al (2023) Created a Large Language Model which was able to simulate Daniel Dennett's writing style and though experts were able to distinguish amongst them at rates above chance, it was surprising how close run the thing was given the fact was that it was scholars who were experts on Dennett who were being probed (ibid).

While a LLM can construct sentences which appear to be analogous to ordinary human sentences there are obviously a lot of disanalogies. While LLMs can reliably produce syntactically sound sentences and sentences which are semantically interpretable. The words the LLM use have no meaning to the LLMs only to the humans that interpret their output. The reason that they have no meaning is because they are not grounded in sensory experience for the LLM. Whereas for humans they obviously are (Harnard 2024). The LLM unlike the human isn't talking about any state of affairs in the world, rather it is merely grouping together tokens according to how the tokens are fed into it in its training data.

When criticisms are made that LLMs outputs don't have meanings. We need to be careful how we parse these statements. Obviously, they have meaning in the sense that they can distinguish between two sentences which are syntactically identical, but which don't have the same meaning. But the sentences do not have meaning in the sense of referential relation between the words and a mind independent reality. However, given that the idea of explicating meaning in terms of a word-world relation has been questioned (Chomsky 2000, Quine 1974), it is difficult to know what to make of the claim that LLMs don't have meanings because their words don't refer to mind independent objects.

Bender & Koller (2020) used a thought experiment to illustrate why they believed that LLMs did not mean anything when they responded to queries. The thought experiment imagines two people trapped on different Islands who are communicating with each other via code through a wire which is stretched between the islands via the ocean floor. In this thought experiment an Octopus who is a statistical genius accesses the wire and can communicate with the other people on the island through pattern recognition. But though he can figure out what code to use, and when, due to the context of the code being used and the patterns of when they are grouped together, he has no understanding of what is being said. Bender & Koller argue that if a person on the Island asked the Octopus how to build a catapult out of coconut and wood he wouldn't know how to answer because he has no real-world knowledge of interacting with the world and is instead merely grouping brute statistical patterns together.

Piantadosi & Hill (2023) in their "Meaning Without Reference in Large Language Models", argue that thought experiments such as the Octopus one fail because it makes the unwarranted assumption that meaning can be explicated in terms of reference. They argue that meaning cannot be explained in terms of reference for the following reasons. There are many terms which are meaningful to us, but which have no clear reference e.g. Justice. We can think of concepts of non-existent objects. These have meanings but don't refer to anything in the mind-independent world. We have concepts of impossible objects such as a round square, perpetual motion machine. We have concepts which pick out nobody, but which are meaningful: e.g. the present King of Ireland. We have concepts which have meaning but which don't refer to concrete particulars e.g. concepts of abstract objects. We have terms which have different meanings but the same reference e.g. morning star-evening star (Piantadosi & Hill 2023).

They go on to argue that conceptual role theory in which meaning is determined in terms of entire structured domains (like Quine's web-of-belief) plays a large role in our overall theory of the world. But they do nonetheless acknowledge that reference plays some role in grounding our concepts. Just not as large a role as some theorists criticizing LLMs believe. They are surely right that as theory of the world becomes more and more sophisticated our theory will, as Quine noted, face empirical checkpoints only at the periphery (Quine 1951). Nonetheless, when humans are acquiring their language in childhood they must go through a period where they learn to use the right word in the presence of the right object, and to somehow learn to triangulate with their peers in using the same word to pick out a common object in their environment.

As we saw above Piantadosi & Hill (2023) shared concerns about crude referentialist theories of meaning and their relation to LLMs, but they did acknowledge that there are word-world connections between some words and objects in the mind independent world (Ibid.). Quine famously tried to connect our sentences to the world through his notion of an observation sentences (Quine 1960). He argued that an observation sentence was a sentence which members of a verbal community would immediately assent to in the presence of the relevant non-verbal stimuli (ibid).

But he immediately ran into a difficulty with this approach. The difficulty stemmed from the fact that he found it hard to say why different members of a speech community assented to an observation sentence. He tried to cash out the meaning in terms of stimulus meaning where a particular observation sentence was associated with a particular pattern of sensory receptors being triggered. But this made intersubjective assent on observation sentences difficult to explain given that different subjects would obviously have different patterns of sensory receptors triggered in various ways in response to the same observation sentences. What pattern of sensory receptors was triggered by what observation sentences would be largely the result of each subject's long forgotten learning history (Quine 1990). All of this made it difficult to see how Quine could make sense of a community assenting to an observation sentence being used in various circumstances.

Quine eventually made sense of intersubjective assenting to an observation sentence by appealing to the theory of evolution. He argued that there was a pre-established harmony between our subjective standards of perceptual similarity and trends in the environment (Quine 1996). And that humans as a species were shaped by natural selection to ensure that they shared perceptual similarity standards (Ibid). This fact was what made it possible for humans to share assent and dissent to observation sentences being used in certain circumstances (Ibid).

For Quine shared perceptual similarity standards and reinforcement for using certain sounds in certain circumstances gave observation sentences empirical content. But to achieve actual reference Quine argues that we need to add things like quantifiers, pronouns, demonstratives etc (Quine 1960). The key point is though that observation sentences link with the world (even if in a manner less tight than objective reference), because of our shared perceptual similarity standards matching objective trends in our environment. This is what makes intersubjective communication possible. LLMs are not responsive to the environment in anything like the way humans are when they are acquiring their first words. Even prior to humans learning the referential capacities of language, children using observation sentences are still in contact with the world. To this degree then Piantadosi & Hill's concerns about reference are besides the point. As humans begin to acquire their first words, they do so through observation sentences which are connected to our environment. The fact that when we acquire a language complete with words and productive syntax we can speak about theoretical items, fictional items, impossible objects etc is interesting but doesn't speak to the LLM issue. When Humans first acquire their words, they do so in response to their shared sensory environment, LLMs do not learn in this way at all. Their training is entirely the result of exposure to textual examples which they group into tokens based on the statistical likelihood of textual data occurring together. So, while humans eventually learn to speak about non sensorially experienced things their first words are keyed to sensory experience, and this is a key difference between them and LLMs.

## Some Push Back

The idea that humans and LLMs differ because humans are referentially grounded and LLMs are not has been challenged by several theorists (Sogaard 2022, Molo & Milliere 2023, Mandelkern & Linzen 2024). One of the key points they make is that LLMs have outputs which are isomorphic with aspects of the mind independent world. Thus Sogaard 2022 makes the following claim:

> *"Words that are used together, tend to refer to things that, in our experience, occur together. When you tell someone about your recent hiking trip, you are likely to use words like mountain, trail, or camping. Such words, as a consequence, end up close in the vector space of a language model, while being also intimately connected in our mental representations of the world. If we accept the idea that our mental organization maps (is approximately isomorphic to) the structure of the world, the world-model isomorphism follows straight-forwardly (by closure of isomorphisms) from the distributional hypothesis."* (Sogaard 2022 p.443).

While it is acknowledged by all that mere isomorphism alone obviously isn't sufficient to provide referential grounding. Some theorists argue that there are other factors which can be appealed to along with isomorphism which show that LLMs are referentially grounded.

Molo & Milliere (2023) have claimed that there is a degree of referential grounding the output of LLMS can attain. They aren't arguing that the LLMs themselves ground the symbols, rather they claim that the outputs themselves are grounded. They then note that there are two key criteria necessary to achieve referential grounding (1) causal informational relations between representations and the world, (2) Historical relations which imbue the systems with a type of normativity. And they maintain that with LLMs 1 is provided by human linguistic structures partially mirroring reality, and 2 is provided by reinforcement learning, specifically, that part of the training which involve training to represent reality accurately.

When discussing causal informational relations between LLMs outputs and the world they make the following claim:

> *"…we should expect LLMs that learn to encode word meanings based on co-occurrence statistics to acquire linguistic representations whose structure in the high-dimensional vector space can be mapped to the structure of the world, at least in limited domains."* (Molo & Milliere 2023 p. 18).

They justified this claim by appealing to recent experimental work which demonstrated isomorphisms between LLM outputs and spatial features of the world in areas such as absolute spatial representations of cities, which cities shared boarders, the geometry of colour outputs etc (ibid p. 18). These results do indicate a causal informational relational relation between LLMs and the world which is to be expected given that the LLMs are trained on human linguistic outputs.

Molo & Milliere then argue that the LLMs meet the normative criteria because LLMs are trained using reinforcement training. This reinforcement training has two cores an ethical core and an epistemic core. The training consists in rewarding outputs that meets ethical standards using a rating system, and punishing outputs that don't meet the standards by rating them low. And similarly, there is epistemic training which rewards and punishes outputs depending on how accurately they mirror reality.

The argument being that since the outputs share a causal informational relation with the mind independent world, and they are trained from a normative perspective to ensure this relation is in place then we can say that the outputs are referential.

To be clear nobody is claiming that the LLMs are using words to refer. Rather what is at issue is whether the outputs themselves are referential. Mandelkern & Linzen (2024) draw a contrast between two types of outputs which illustrates the point nicely. They contrast an ant who accidently draws an outline of the sentence "Peano proved the incompleteness Theorem" in the sand and a first-year logic student called Luke who sends a text saying "Peano proved the incompleteness theorem". They then ask whether a LLM which created an output of the same sentence would be closer to an ant or a human who had sent the text (ibid p.2). In the human case we would say that he is referring to Peano and saying something false about him. While with the Ant we would say he is referring to nothing at all.

Luke our first-year logic student doesn't have to have any complex beliefs and experiences in relation to Peano. But when he speaks about "Peano" he is referring to a person that we can trace back to an original baptism event where the name was given to Peano. There is a causal historical relation between this original baptism event and the name as we originally use it. Typically, it is claimed that LLMs aren't grounded because they are trained entirely on text. But Mandelkern & Linzen argue that in the case of Luke he had no interaction with Peano at all, all he knows is the sentence he incorrectly learned about him, and the sentence refers because of its causal historical relation to the original object; Peano. Being part of our speech community means that Luke can refer to Peano despite knowing little about him and never interacting with him. And they argue that since LLMs are trained on human texts which can be causally-historically traced back to the original baptism we can say that if LLMs are part of our speech community it's terms are referential. Hence, they argue a LLM is closer to the Logic student Luke than it is to the Ant accidently tracing a sentence in the sand.

The case for referential grounding for LLMs is threefold. (1) They are trained on texts which are causally historically related to original baptism events giving some of their outputs referential relations. (2) LLM outputs because of their training on human text which was originally used to refer to the world have isomorphic features in common with the world. (3) There is a normative element to training which emphasises accuracy of depiction of reality. This is analogous to the way that children learn words referring to objects they have no experience of because the words they acquire have a causal historical relation to the original baptism event. They learn facts in books which are isomorphic to facts in the objective world even if they have no experiential relation to these facts (e.g. The Earth is the third planet from the Sun). And they have normative systems in place which will reinforce them for using accurate information and punish them for using inaccurate information.

**Replying to the Push-Back**

Originally, we have noted that much of human language is not referential hence arguments that LLMs are not referentially grounded aren't as impactful as sometimes believed. But it was noted via a discussion of Quine's notion of stimulus meaning that nonetheless humans differ from LLM because humans learn their first words through intersubjectively keying them to stimulus meanings, and LLMs are not capable of doing this. But in the last section we considered the rejoinder that LLMs are referentially grounded because they have outputs which are causally and informationally related to historical baptismal events, and that their outputs are shaped by normative considerations emphasizing accuracy which is implemented through reinforcement learning.

But Quine's notion of stimulus meaning is more fundamental than referential relations. Quine correctly notes that when we are learning the stimulus meaning we do so before we acquire the capacity to use objective reference:

> *"But the mother, red, and water are for the infant all of a type: each is just a history of sporadic encounter, a scattered proportion of what goes on. His first learning of the three words is uniformly a matter of learning how much of what goes on about him counts as the mother, or as red or as water. It is not for the child to say in the first case "Hello Mama! Again," in the second case "Hello! another red thing," and in the third case "Hello! More water. They are all on a par: Hello! more mamma, more red, more water."* (Quine 1969 p. 7).

On Quine's conception the child hasn't yet learned the syntax of quantification which requires using quantifiers, relative terms, pronouns etc. Therefore, the child is not capable of making ontological distinctions which would be necessary for the child to use objective reference prior acquiring this developmental capacity[8]. Nonetheless the child is still associating the sounds with stimulus meaning derived from his perceptual environment. And here we can see that the human and the LLM are entirely distinct. Even if we grant the externalist his argument that the LLM is capable of referential grounding the LLM unlike the human is not engaging in sensorial grounding. Hence, the competency that the LLM is exhibiting is very different than the competency that the human exhibits.

### AI and the Relevance of Rationalism and Empiricism.

Above we discussed some disanalogies between LLMs and human linguistic capacities. Two main differences were noted (1) Differences in the stimulus needed for the respective agents to acquire language, which indicates different underlying competencies. (2) LLMs language are not sensorily grounded, and human language is sensorily grounded; and this difference is a result of the different ways in which they acquire their language, humans begin by being responsive to the world in a triangular relation with others, while the LLM acquires their language through statistical grouping of text they are trained on.

---

[8] Carey 2009 amasses an impressive amount of experimental data that indicates that Quine was wrong. Children can make ontological distinctions prior to grasping the syntax of quantification. But this data is fatal for proponents of the view that LLMs are sensorily grounded because it shows the incredible cognitive complexity necessary for a child to learn to learn to refer to objects. This cognitive complexity is entirely absent from LLMs (See Marcus 2020).

So given that humans and LLM's outputs are the result exposure to different quantities, and types of data, which indicate different underlying competencies, a question arises as to the relevance of AI to understanding human linguistic capacities. Earlier we discussed the debate between Chomsky and Skinner & Quine and its relation to current debates on the nature of LLMs. But given the disanalogies we have noted between LLMs and humans it is questionable whether they have anything to tell us about the rationalist-empiricist debate at all.

The debate between the rationalists and empiricists was never centred on whether innate apparatus was necessary for a creature to learn a particular competency. All sides of the debate agreed that some innate apparatus was necessary to explain competencies, and the degree of the innate apparatus which need to be postulated is to be determined empirically. The empiricist position was one which argued that humans learned primarily through data driven learning (supported by innate architecture), while the rationalist argued that humans learned through innate domain specific competencies being triggered by environmental input.

However, given the differences between LLMs and Human's linguistic competencies their relevance to each other on the rationalist-empiricist debate is in doubt. Even if we can conclusively demonstrate that a LLM learns in an empiricist manner, this will not tell us anything about whether humans learn in an empiricist manner. Because human competencies are so different than a LLMs it is simply irrelevant to the rationalist-empiricist debate whether LLMs learn in an empiricist manner or not.

It is theoretically possible that a philosopher could argue a la Kant that any form of cognition will a priori need to implement innate domain specific machinery to arrive at its steady state. And one could offer empirical data to support this a priori claim by showing that very different forms of cognition e.g. human and LLMs both learn by implementing innate domain specific architecture. But this isn't how the debate has been played out in the literature. Typically, the literature argues that LLMs are largely empiricist, and this fact vindicates empiricism in general, or it is argued that they are largely rationalist, and this fact vindicated rationalism. I have argued here that given the very different nature of LLMs and humans it is irrelevant to the question of how humans acquire their knowledge whether LLMs are rationalist or empiricist.

But this isn't to say that it is unimportant whether LLMs or other forms of AI learn in a rationalist or an empiricist manner. There are still practical issues in engineering as to whether one is more likely to be successful in building things like Artificial General Intelligence using empiricist architecture or not. Thus, people like Marcus (2020) argue that while we *may* be able to build AGI using empiricist principles, we will not be able to build AGI unless we build in substantial innate domain specific knowledge into the system. Marcus even argues that the best way to understand what innate architecture is necessary to be built into our AI models we should look to our best example of an organism with general intelligence, i.e. Humans.

While the engineering question is extremely interesting from a practical point of view and could motivate an interest in whether LLMs or other types of AI learn in a rationalist or an empiricist manner. But when it comes LLMs, the question of whether they learn in an empiricist, or a rationalist manner is largely irrelevant to the rationalist-empiricist question in relation to humans.

## Conclusion

This paper evaluated the rationalist empiricist debate between Chomsky, Quine and Skinner and its relation to current debates about the architecture used by LLMs. It showed that LLMs and Humans learn differently indicating different underlying competencies. It therefore argued that whether LLMs learn in a rationalist, or an empiricist manner is irrelevant to whether humans are rationalist or empiricist learners. The paper concluded by arguing that since humans are the paradigm example of general intelligence learners. If engineers want to build AGI, they should do so by modelling it on human cognition. But doing this involves sticking as close as possible to the science of human cognition and behaviour and moving away from arcane debates such as rationalism vs empiricism.

## Bibliography

Baggio, G & Murphy, E. 2024. "On the Referential Capacities of Language Models: An Internalist Rejoinder to Mandelkern & Linzen". https://arxiv.org/abs/2406.00159

Berwick, R, & Pietroski, P, & Chomsky, N. 2011 "Poverty of the Stimulus Revisited." *Cognitive Science. Vol 35. Issue 7 pp. 1207-1242.*

Bender & Koller. (2020) "Climbing Towards NLU: On Meaning, Form and Understanding in the Age of Data." *Proceedings of the 58th annual meeting of the Association for Computational Linguistics.* https://aclanthology.org/2020.acl-main.463/

Boeckx, C (2006). *Linguistic Minimalism.* Oxford. Oxford University Press.

Brown, R. & Hanlon, C. 1970. "Derivational complexity and order of acquisition in child speech." In Hayes, J.R. (eds). *Cognition and the Development of Language.* New York Wiley.

Buckner, C. 2018. "Empiricism without Magic: Transformational Abstraction in Deep Convolutional Neural Networks. *Synthese. Vol 195 pp. 5339-5372.*

Carey, S. 2019. *The Origin of Concepts.* Oxford University Press.

Childers, T, & Hvorecky, J, & Majer, O. 2023. "Empiricism and the foundations of Cognition". *AI and Society. Vol 38 pp. 67-87.*

Chomsky, N. 1959. "A review of B.F. Skinner's Verbal Behaviour". *Language, 35. PP 26-57.*

Chomsky, N. 1965. *Aspects of a Theory of Syntax.* MA: MIT Press.

Chomsky, N. 1966. *Cartesian Linguistics.* New York: Harper & Row.

Chomsky, N. 1969. "Quine's Empirical Assumptions". *Synthese 19 pp. 53-68.*

Chomsky, N. 1975. *Reflections on Language.* New York: Random House.

Chomsky, N. 1986. *Knowledge of Language: Its Nature, Origin and Use.* New York, New York: Praeger.

Chomsky, N. 2000. *New Horizons in the Study of Language and Mind*. Cambridge: MA.

Chomsky, N, Roberts, I, Watumull, J. 2023. "The False Promise of ChatGPT." *The New York Times.*

Chomsky, N & Katz, J. 1975. "On Innateness: A Reply to Cooper." *Philosophical Review. 84 pp. 70-84.*

Clark, L & Lappin, S. 2011. *Linguistic Nativism and the Poverty of the Stimulus.* Wiley

Collins, J. 2007. "Linguistic Competence Without Knowledge of Language". *Philosophy Compass 2 (6) pp. 880-895*

17Crain, S & M, Nakayama. 1987. "Structure Dependence in Grammar Formation". *Language, Vol 63 pp. 522-543*.

Firestone, C. (2022) "Competence and Performance in Human Machine Comparisons." *Proceedings of the National Academy of Sciences. 117. 43. Pp. 25662-26571*.

Fodor, J. 2003. *Hume Variations*. Oxford University Press.

Fodor, J & Pylyshyn 2015. *Minds without Meanings: An Essay on the Content of Concepts*. The MIT Press.

Hart, B & Risley, T, & Kirby, J. 1997. "*Meaningful differences in the everyday experiences of the young American.*" *Canadian Journal of Education. Vol 22. Issue 3*.

Harnard, S. 2024. "Language Writ Large: LLMs, ChatGPT, Grounding, Meaning, and Understanding." https://arxiv.org/abs/2402.02243

Hornstein, N 2005. "Empiricism and rationalism as research strategies". *The Cambridge Companion to Chomsky pp 145-163*.

Jackendoff, R. 2002. *Foundations of Language.* Great Clarendon Street. Oxford University Press.

Jennings, Adrianne, M. et al. (2021). "A systematic review of empirical Intraverbal Research: 2015-2020." *Behavioural Interventions. 37. 1. Pp. 79-104*.

Katzir, R. 2023 "Are Large Language Modules Poor Theories of Linguistic Cognition: A Reply to Piantadosi". *Biolinguistics. 17: pp. 1-12*.

Kodner, J, & Payne, S, & Heinz, J. 2023 "Why Linguistics will thrive in the 21th Century: a reply to Piantadosi". https://arxiv.org/abs/2308.03228

Laka, I. 2009 "What is there in Universal Grammar: On innate and specific aspects of Language. In Piattelli-Palmarini, M, Uriagereka, J, & Pello Salaburo (EDS) *Of Minds and Language*. Oxford. Oxford University Press.

Long, R. 2024. "Nativism and Empiricism in Artificial Intelligence". *Philosophical Studies. Vol 181 pp 763-788*.

Lyons, J. 1970. *Chomsky*. Collins, Fontana.

MacCorquodale, K. 1970. "On Chomsky's Review of Skinner's Verbal Behaviour". *Journal of the experimental analysis of behavior. 13. 1. p. 83*.

Mandelkern, M & Linzen, T. 2024. Do Large Language Models Refer? *Computational Linguistics*. Pp. 1-10.

Marcus, G. 2020. "The Next Decade in AI: Four Steps Towards Robust Artificial Intelligence"

Mollo, D, & Milliere, R. 2023. "The Vector Grounding Problem". https://arxiv.org/abs/2304.01481

Milliere, R. 2024. "Language Models as a Model of Language". In Nefdt, R., & Dupre, G., & Stanton K. (eds) *The Oxford Handbook of the Philosophy of Linguistics.* Oxford University Press.

Milliere, R & Buckner C. 2024a. "A Philosophical Introduction to Language Models: Part 1". https://arxiv.org/abs/2401.03910

Milliere, R & Buckner C. 2024b. "A Philosophical Introduction to Language Models: Part 2".

https://arxiv.org/abs/2405.03207

Pavlick, E. 2023 "Symbols and Grounding in Large Language Models". *Philosophical Transactions of the Royal Society A,* 381 (2251). 20220041.

Petursdottir, A, I. & Devine, B. 2017. "The Impact of Verbal Behaviour on the scholarly literature from 2005 to 2016. *The Analysis of Verbal Behaviour, 33. Pp. 212-228.*

Piantadosi, S. 2023 "Modern Language Models Refute Chomsky's Approach to Language."

https://lingbuzz.net/lingbuzz/007180/v6.pdf

Piantadosi, S, & Hill, F. 2023 "Meaning with our Reference in Large Language Models".

https://arxiv.org/abs/2208.02957

Pinker, S. 2002. *The Blank Slate: The Modern Denial of Human Nature.* New York Viking.

Pullum, G, K. & Scholz, B. 2002 "Empirical assessment of stimulus poverty argument." *The Linguistic Review. 19 pp. 9-50.*

Quine, W. 1960. *Word & Object.* The MIT Press. Cambridge MA.

Quine, W. 1968. "Reply to Chomsky". *Synthese* 19 pp. 274-283.

Quine, W. 1969. *Ontological Relativity and Other Essays.* Cambridge University Press. New York.

Quine, W. 1970. "Methodological Reflections on Current Linguistic Theory". *Synthese 19, pp. 264-321.*

Quine, W. 1974. *The Roots of Reference.* La Salle. Open Court Press.

Quine, W. 1976. *The Ways of Paradox and other essays.* Cambridge MA. Harvard University Press.

Quine, W. 1990. *Pursuit of Truth.* Cambridge MA. Harvard University Press.

Quine, W. 1996 "Progress on Two Fronts". In Follesdal and Quine 2008 (Eds) *Quine: Confessions of a Confirmed Extensionalist and Other Essays.* Harvard University Press.

Sampson, G. 2002. "Exploring the Richness of the Stimulus". *The Linguistic Review.* 19 pp. 73-104.

Schwitzgebel, E, & Schwitzgebel, D, & Strasser, A. 2024. "Creating a Language Model of a Philosopher". *Mind & Language. 32 (2) 237-259.*

Skinner, B.F. 1957. *Verbal Behaviour.* New Jersey: Prentice-Hall Inc.

Smith, B. 2014. "Quine and Chomsky on the ins and outs of Language." In Harman, G., & Lepore, L. (EDS). *A Companion to WVO Quine.* Wiley & Sons.

Smith, L, D. 1986. *Behaviourism and Logical Positivism: A Reassessment of the Alliance.* Stanford University Press.

Sogaard, A. 2022. "Understanding Models, Understanding Language." *Synthese.* 200 (6) 443.

Wilson, K, G, O Donohue, W, & Hayes, S. (2001) "Humes Psychology, contemporary learning theory, and the problem of knowledge amplification," *New Ideas in Psychology. 19.1. pp. 1-25.*

["